%% file: iclr2027_conference.tex
\documentclass{article} 
\usepackage{iclr2027_conference,times}

\input{math_commands.tex}

\usepackage[utf8]{inputenc} 
\usepackage[T1]{fontenc}    
\usepackage{hyperref}       
\usepackage{url}            
\usepackage{booktabs}       
\usepackage{amsfonts}       
\usepackage{nicefrac}       
\usepackage{microtype}      
\usepackage[table]{xcolor}         
\definecolor{ForestGreen}{RGB}{34,139,34}
\definecolor{FigOneGreen}{RGB}{48,110,28}

\usepackage{amsmath}
\usepackage{cleveref}
\usepackage{graphicx}
\usepackage{caption}
\usepackage{tcolorbox}
\usepackage{algorithm}
\usepackage{algpseudocodex}
\usepackage{makecell}
\usepackage{inconsolata}
\usepackage{nicematrix}
\usepackage{tabularx}
\usepackage{wrapfig}

\usepackage{fontawesome5}

\usepackage{overpic}
\usepackage{tikz}

\newcommand\encircle[1]{%
  \tikz[baseline=(X.base)]
    \node (X) [draw, shape=circle, fill=black, inner sep=-1pt] {\strut\resizebox{.75\width}{!}{\textcolor{white}#1}};}
\newcommand\encirclefigone[1]{%
  \tikz[baseline=(X.base)]
    \node (X) [shape=circle,fill=FigOneGreen,inner sep=-1pt] {\strut\resizebox{.75\width}{!}{\textcolor{white}#1}};}
\newcommand*\captionlabelsmall[1]{%
  \tikz[baseline=(char.base)]{%
    \node[shape=circle,fill=black,draw=black,inner sep=1.5pt, outer sep=5pt] (char) {%
      \textcolor{white}{\tiny \textsf{#1}}};
}}

\title{ExperienceIndex: Artifact-Grounded Memory}

\iclrfinalcopy

\author{%
Peter Baile Chen$^{1}$ \quad Geoffrey X. Yu$^1$ \quad Xinming Liu$^2$ \quad
\textbf{Samuel Madden}$^1$ \\
\textbf{Dan Roth}$^3$ \quad  \textbf{Jacob Andreas}$^1$ \quad \textbf{Doug Downey}$^4$ \quad \textbf{Michael Cafarella}$^1$\\
$^1$MIT \quad $^2$McKinsey \& Company \quad $^3$Oracle AI \& UPenn \quad $^4$AI2\\
\small{Correspondence: \texttt{peterbc@mit.edu}}\\
}

\newcommand{\sys}{\texttt{ExpIdx}}
\newcommand{\reasoningbank}{\texttt{ReasoningBank}}
\newcommand{\enrichindex}{\texttt{EnrichIndex}}

\newcommand{\BestAnswerQualityImprovement}{11.0}
\newcommand{\BestCostReduction}{50.5\%}

\begin{document}

\maketitle
\lhead{}

\input{sections/abstract}
\input{sections/intro}
\input{sections/related}

\input{sections/method}

\input{sections/exp}

\input{sections/analysis}
\input{sections/conclusion}

\subsection*{AI use statement}

In this work, we used generative AI tools for proofreading the paper and to
assist in implementing parts of the \sys{} prototype.
We have not used generative AI tools for any other tasks with required disclosure.
We have reviewed all AI-assisted work.
We checked and manually integrated all LLM-suggested proofreading changes.
Additionally, we manually reviewed all LLM generated code and tested it for
correctness by manually checking the program's output on the datasets used in
this paper.

\bibliography{iclr2027_conference}
\bibliographystyle{iclr2027_conference}

\clearpage
\appendix
\input{sections/appendix}

\end{document}

%% file: math_commands.tex
\usepackage{amsmath,amsfonts,bm}

\def\eqref#1{equation~\ref{#1}}

\def\1{\bm{1}}

\DeclareMathAlphabet{\mathsfit}{\encodingdefault}{\sfdefault}{m}{sl}
\SetMathAlphabet{\mathsfit}{bold}{\encodingdefault}{\sfdefault}{bx}{n}



%% file: sections/abstract.tex
\begin{abstract}
Knowledge‑intensive tasks in domains like law, science, and technology require answering many questions by reasoning about a \textit{shared} corpus of artifacts (e.g.\ court cases, scientific literature, or software libraries). As humans interact with these corpora, they naturally accumulate experiential knowledge about artifacts, enabling them to \textit{quickly} identify the \textit{complete} set of \textit{relevant} artifacts for each new task. However, existing AI agents lack appropriate memory solutions to build or reuse such \textit{artifact-grounded experience}, leading to lower answer quality and higher online cost.
Existing memory solutions extract and reuse information from prior task-solving traces, but they primarily focus on user preferences, factual attributes, or abstract reasoning patterns rather than persistent artifact‑specific knowledge.
We introduce \texttt{ExperienceIndex}, a novel experience layer for AI agents that captures and reuses knowledge about artifacts based on prior reasoning traces. \sys{} stores two complementary forms of experience: (i) single‑artifact experiences that summarize an artifact's contribution to prior tasks and (ii) artifact‑pair experiences that encode structural relationships discovered during past reasoning.
Integrated as lightweight middleware, \sys{} uses an experience retrieval mechanism to guide agents toward the \textit{complete} set of \textit{relevant} artifacts for new tasks, improving both answer quality and efficiency.
Across diverse corpora and agentic solutions with different search frameworks, \sys{} delivers consistent gains, raising answer quality by up to \BestAnswerQualityImprovement{} points and reducing online dollar cost by up to \BestCostReduction{}. We further demonstrate two benefits: (i) cross‑task generalization, where experiences accumulated from text‑to‑SQL tasks transfer to factoid QA tasks over the same artifact corpus, and (ii) teacher-student learning, where experiences from a stronger model enable a weaker model to reach comparable performance.
These results highlight \sys{} as a general, modular approach for enabling agents to build and leverage persistent, artifact‑grounded experience.
\end{abstract}

%
%

%% file: sections/intro.tex
\section{Introduction}
\label{sec:intro}


\begin{figure}
\centering
\begin{overpic}[width=0.99\textwidth, clip, trim={0cm 0.9in 0cm 0cm}]{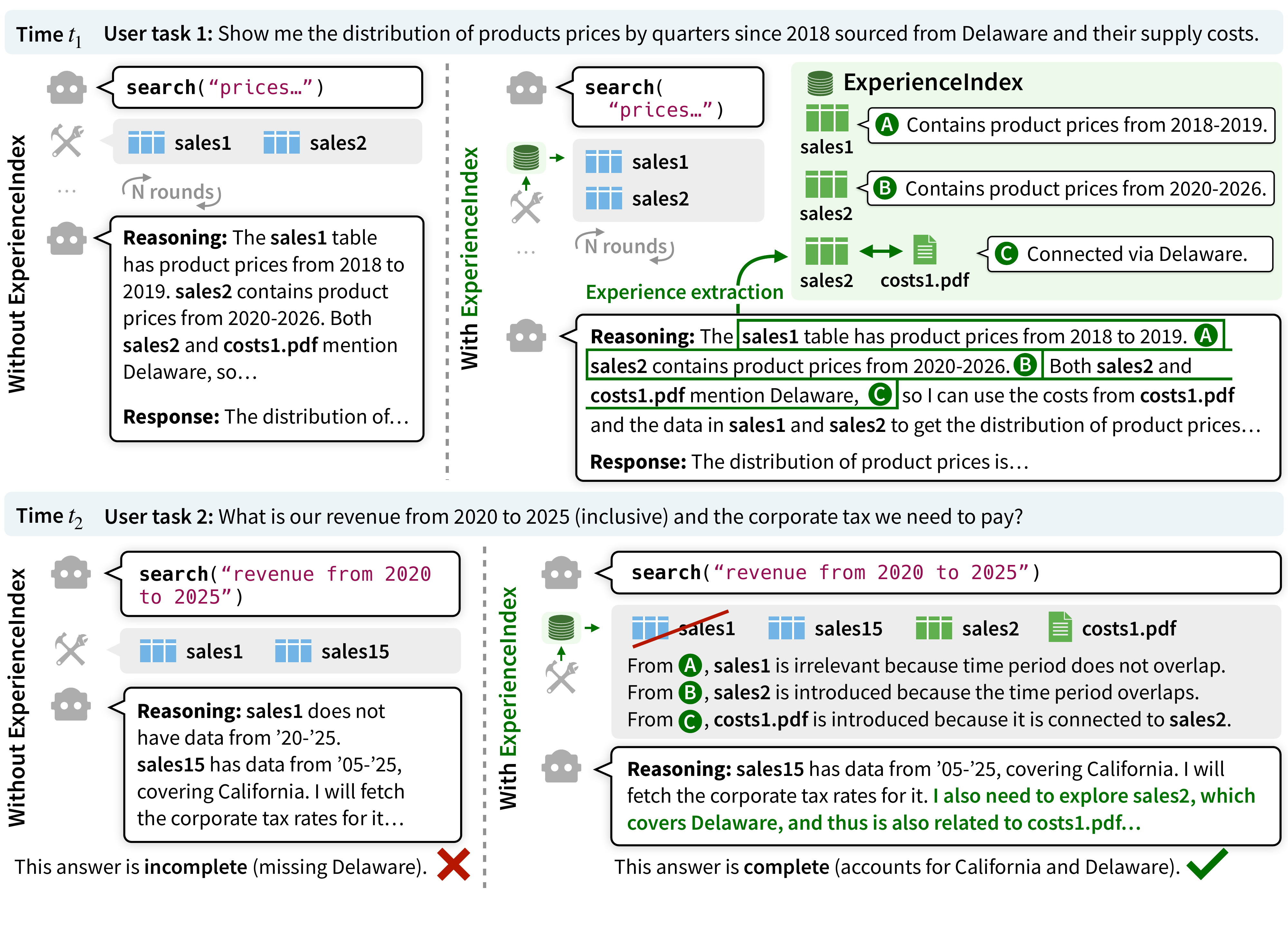}
\end{overpic}
\vspace{-0.5em}
\caption{%
\sys{} (green) and its integration into an agentic solution as middleware.
Without \sys{}, an agentic solution iteratively explores the artifact corpus for \textit{each} user task using search tools, possibly in many rounds, to gather the complete information needed to solve the task (an error‑prone and inefficient process).
\sys{} optimizes this process by giving the agent the full set of relevant artifacts derived from past experiences, improving answer quality while reducing execution cost.
}
\label{fig:diagram}
\vspace{-1em}
\end{figure}

Knowledge-intensive fields, such as law, consulting, technology, and science, require humans to solve many different tasks using a \textit{shared} corpus of artifacts, including documents, tables, files, and codebases. For humans, repeated interaction with this corpus does more than resolve each individual task: it gradually builds familiarity with the corpus itself. The first encounter is slow and labor‑intensive, involving extensive reading and tracing cross‑references to figure out which artifacts matter. Over time, however, repeated interactions leave behind conceptual anchors tied to specific artifacts: mental notes or literal sticky notes like ``this table contains revenue from 2015–2025'' or ``ignore that tax document; it does not mention Delaware.'' When new tasks appear, humans draw on this accumulated \textit{artifact-grounded experience} to \textit{quickly} identify the \textit{complete} set of \textit{relevant} information, improving both answer quality and efficiency.

While humans naturally accumulate experiential knowledge about artifacts that they have previously examined, today's agentic systems rely on external memory mechanisms to achieve statefulness across tasks and sessions.
However, existing memory mechanisms are limited: they
store user preferences or facts~\citep{openai_memory_faq, chhikara2025mem0} or rely on ad‑hoc scratchpad mechanisms~\citep{claude_memory_tool}.
Several \textit{abstracted-memory} mechanisms have also been proposed~\citep{suzgun2026dynamic, chen2025log, ouyang2025reasoningbank}, focusing primarily on extracting and reusing generalizable reasoning patterns. For example, \cite{suzgun2026dynamic} extracts key solution strategies from reasoning traces of past math problems to improve accuracy:
\textit{``1. State problem requirement clearly 2. List key [...] theorems applicable...''}.
Similarly, \cite{ouyang2025reasoningbank} extracts key web navigation strategies from past reasoning traces to improve efficiency: \textit{``Use filter buttons instead of scrolling to navigate to products of certain categories''}.
However, these approaches still do not address the challenge of retaining and reusing artifact‑specific knowledge.

We introduce \texttt{ExperienceIndex} (abbreviated as \sys{}), a novel experience layer for AI agents that accumulates analysis from prior tasks (task 1 in the example illustrated in \Cref{fig:diagram}) and grounds these experiences in the specific artifacts involved (\texttt{sales1}, \texttt{sales2}, \texttt{costs1.pdf}).
Concretely, experiences are represented through (i) single-artifact experiences that capture how specific artifacts contribute to solving prior tasks (\encirclefigone{A} and~\encirclefigone{B})
and (ii) artifact-pair experiences that encode previously discovered structural connections between artifacts (\encirclefigone{C}). \sys{} then acts as a middleware between past experiences and model context that aims to use experience retrieval to filter out irrelevant artifacts (\texttt{sales1}) and identify the full set of relevant ones (\texttt{sales2} and \texttt{costs1.pdf}) for future tasks (task~2), improving answer quality while reducing computational cost.

Extending from the example in \Cref{fig:diagram}, we highlight two key dimensions of \sys{}'s generality. First, \sys{} stores experiences over single artifacts or artifact pairs, independent of modality, which allows it to support a wide range of real‑world workloads operating over a \textit{shared} corpus, including conversation logs, code repositories, scientific literature, and relational tables.
Second, \sys{} functions as a lightweight middleware layer, enabling it to integrate with diverse search frameworks across different solutions, such as pipelined retrieval components, retriever-based tools, or file-system-based agentic search.
To validate this generality, we evaluated \sys{} across a broad suite of knowledge‑intensive tasks and heterogeneous search frameworks (\Cref{tab:dataset}).
Across all settings, relative to existing agentic solutions, \sys{} consistently improves answer quality by up to \BestAnswerQualityImprovement{} points while reducing online cost by up to \BestCostReduction{} (\Cref{tab:perf-1}).
We further observe additional benefits of our solution: (i)~\textit{cross‑task generalization} where experiences populated by \sys{} from text‑to‑SQL tasks can also benefit factoid QA tasks over the same table corpus (\Cref{tab:perf-1}), and (ii)~\textit{teacher-student learning} where experiences from a stronger model allow a weaker model to reach comparable performance (\Cref{tab:perf-3}).

%% file: sections/related.tex
\section{Related work}
\label{sec:related}


\textbf{Long-term memory for AI agents.}
Existing long‑term memory solutions store information that can be retrieved to support reasoning on future tasks. Broadly, they fall into two categories: \textit{profile‑based memory} and \textit{abstracted memory}.
Profile‑based memory solutions~\citep{openai_memory_faq,chhikara2025mem0} use the \textit{user profile} as the anchor, storing preferences (e.g., seat choices, writing style), conversation history, and factual attributes (e.g., birthday, address).
\cite{shu2026remem} extends this by structuring user information along a temporal axis. However, these solutions primarily target tasks over conversation history~\citep{maharana2024evaluating}.
In contrast, \sys{} treats \textit{artifacts}, not users, as anchors, and stores experience grounded in those artifacts, making it applicable not only to conversation logs, but also to a wide range of workloads involving a shared corpus.
A second line of work focuses on abstracted memory~\citep{suzgun2026dynamic,chen2025log,ouyang2025reasoningbank}, introduced in \Cref{sec:intro}, which targets tasks that share underlying \textit{reasoning patterns or computations}.
\sys{}, by contrast, targets tasks that share \textit{knowledge about artifacts}. Rather than extracting general logic, it captures analyses, contributions, and relationships among artifacts from prior reasoning traces. We view our work and abstracted memory as complementary components needed for solving complex tasks.


\textbf{Offline corpus enrichment.}
Prior work~\citep{chen2025enrichindex,edge2024local,gospodinov2023doc2query,gutierrez2024hipporag} has explored enriching corpus artifacts offline to enhance retrieval quality. These approaches augment artifacts with annotations, such as layperson summaries, synthetic question-answer pairs, or graph-structured representations.
But these methods have limitations:
(i)~the annotations are generated without visibility into downstream tasks, so their benefit for real retrieval needs may be limited;
(ii)~they incur substantial up‑front cost, since every artifact (or combinations of artifacts) must be processed in advance; and
(iii)~they assume access to the entire corpus during offline indexing (often unrealistic, e.g., for open-web content).
In contrast, \sys{} updates its memory store continually and incrementally, based solely on artifacts that appear in the prior reasoning traces of executed tasks. This design allows \sys{} to 
(i) evolve in response to real task demands, (ii) avoid up-front cost, and (iii) operate without assuming global corpus visibility.


%% file: sections/method.tex
\section{ExperienceIndex}
\label{sec:expindex}




This work focuses on tasks involving reasoning over a shared corpus of artifacts, structured as pairs of <\texttt{artifact\_id}, \texttt{artifact\_content}> along with optional metadata (with \texttt{artifact\_id} auto-populated when absent).
Today's AI agents typically approach such tasks using a ReAct loop~\citep{yao2022react} built on top of a solver model $\mathcal{M}$ and a search framework $\mathcal{R}$ comprising tools (e.g., \texttt{vector\_search}, \texttt{bash}).
The agent iteratively explores the corpus to retrieve information needed to solve the task (see \Cref{fig:diagram}, left half).
However, as discussed in \Cref{sec:intro,sec:related}, existing memory solutions fail to capture and reuse experiential knowledge about artifacts the agent has previously examined. As a result, they miss opportunities to accelerate and improve retrieval and reasoning on future tasks.
To address this gap, we introduce \sys{}, which treats artifacts as anchors for persistent memory.

\Cref{alg:expindex} illustrates how \sys{} integrates with the agentic setup described earlier. At its core is a memory store that accumulates experiences extracted from past reasoning traces, which are indexed by \texttt{artifact\_id}s.
When a new task arrives, \sys{} acts as middleware between the memory store and the model context: it retrieves relevant past experiences based on the task description (line 4) or on the set of \texttt{artifact\_id}s (line 10) returned by the search framework (line 9), and injects them into the model context (lines 5 and 11). After the agent completes its reasoning, \sys{} extracts and stores experiences about the artifacts used in the reasoning trace (line 13).
Because \sys{} derives knowledge about artifacts from the model's reasoning traces, the types of artifacts it supports align with $\mathcal{M}$'s modalities.
If $\mathcal{M}$ is text-only, \sys{} supports text-based artifacts (e.g., text documents, code files) or artifacts that can be transcribed into text. If $\mathcal{M}$ is multi-modal and supports images or other modalities, \sys{} can likewise support artifacts in those formats.
Additionally, \sys{} is compatible with different search frameworks $\mathcal{R}$ whose outputs include \texttt{artifact\_id}s (line 9), enabling retrieval of relevant past experiences as detailed above.

To enable the above process over diverse corpora, \sys{} must (i) store experiences in a domain-agnostic representation (\Cref{sec:expindex-representation}), (ii) retrieve experiences relevant to a new task (\Cref{sec:expindex-retrieval}), (iii) inject them into the model context (\Cref{sec:expindex-injection}), and (iv) extract and index new experiences from the model's reasoning traces (\Cref{app:expindex-indexing}).

\begin{algorithm}
\small
\caption{\sys{} integrated into a standard agentic solution with a solver model $\mathcal{M}$, and a search framework $\mathcal{R}$ (e.g., \texttt{vector\_search)} over a corpus $\mathcal{C}$ of artifacts $\mathcal{A}$ ($\mathcal{A}\textsubscript{ID}$ refers to the \texttt{artifact\_id}s). 
\textcolor{ForestGreen}{Green text shows additions introduced by \sys{}.}}
\label{alg:expindex}
\begin{algorithmic}[1]
\Require Solver model $\mathcal{M}$, search tool $\mathcal{R}$, corpus $\mathcal{C}$, system prompt $\mathcal{P}$, \textcolor{ForestGreen}{memory store $\mathcal{S}$}
\State \textbf{Input:} Input task $t$, \textbf{Output:} An answer
\vspace{0.35em}
\Function{RetrieveExperience}{$\mathcal{S}$: Memory store, $q$: Query, $\mathcal{A\textsubscript{ID}}$: Artifact IDs}
    \State \textcolor{gray}{Returns relevant experiences and associated artifacts} \Comment{See \Cref{sec:expindex-retrieval}.}
\EndFunction{}
\vspace{0.35em}
\State \textcolor{ForestGreen}{$\mathcal{A}_\mathcal{P} \gets $ \Call{RetrieveExperience}{$\mathcal{S}$, $t$, $\{\}$}} \Comment{Fetch artifacts with experiences relevant to task $t$}
\State \textcolor{ForestGreen}{Inject $\mathcal{A}_\mathcal{P}$ into system prompt $\mathcal{P}$} \Comment{See \Cref{sec:expindex-injection}.}
\State $\mathcal{H} \gets $ Initialize message history with system prompt and other metadata (e.g., tool definitions)
\State $y \gets \mathcal{M}(\mathcal{H})$
\While{$y$ is not the final response}
    \State $\mathcal{A\textsubscript{ID}}$, $\mathcal{A}$ $\leftarrow$ $\mathcal{R}(q, \mathcal{C})$ \Comment{Returns artifacts and their IDs relevant to model-generated query $q$}
    \State \textcolor{ForestGreen}{$\mathcal{A'} \gets $ \Call{RetrieveExperience}{$\mathcal{S}$, $q$, $\mathcal{A\textsubscript{ID}}$}} \Comment{Refine searched artifacts based on past experiences}
    \State Append \textcolor{ForestGreen}{refined tool output $\mathcal{A}'$} to $\mathcal{H}$ \Comment{See \Cref{sec:expindex-injection}.}
    \State $y \gets \mathcal{M}(\mathcal{H})$
\EndWhile
\State \textcolor{ForestGreen}{Enqueue update to $\mathcal{S}$ using reasoning trace extracted from $\mathcal{H}$} \Comment{See \Cref{app:expindex-indexing}.}
\State \Output Model response $y$
\end{algorithmic}
\end{algorithm}



\subsection{Experience representation}
\label{sec:expindex-representation}

A model's success on tasks that require artifact-corpus search depends on two aspects: \textit{relevance} and \textit{completeness} of searched artifacts.
Consequently, \sys{}'s experience representation centers around artifacts. This design (i) ensures that past experiences about specific artifacts can be used to improve both aspects in future tasks, thereby enhancing answer quality, and (ii) allows models to reference prior analyses about artifacts instead of repeatedly performing extensive reasoning from scratch, reducing computational cost.
To this end, we introduce a domain-agnostic representation of \textit{experience} at two complementary levels: single‑artifact experience and artifact‑pair experience.
Single‑artifact experience captures learnings about how individual artifacts helped solve previous tasks (e.g., \encirclefigone{A} and \encirclefigone{B} in the example illustrated in \Cref{fig:diagram}). It is drawn on to assess the \textit{relevance} of an artifact for a future task.
Artifact‑pair experience records previously observed meaningful relationships between artifacts (e.g., \encirclefigone{C}), helping surface related artifacts that
the model might otherwise miss and thereby improve the \textit{completeness} of information.
%



\textbf{Single-artifact experience.}
For each artifact used in prior tasks (e.g., \texttt{sales1}, \texttt{sales2}), \sys{} records
(i) a \textit{task description}, which captures the descriptions of past tasks in which the artifact was used;
(ii) a \textit{contribution summary}, which provides a concise explanation of \textit{how} the artifact contributed to solving each task (e.g., ``\texttt{sales1} contains product prices from 2018-2019'' from~\encirclefigone{A}), and
(iii) \textit{snippets}, which contain the specific portions of the artifact that were used during task solving, optionally augmented with location metadata such as line numbers or offsets to help models quickly locate relevant information.
These three domain-agnostic attributes capture
\textit{how} an artifact was used in past tasks. They provide signals about an artifact's potential relevance (or irrelevance) for future tasks. Furthermore, snippets capture only the most pertinent content, which helps models avoid reprocessing large amounts of irrelevant material from the same artifact.

\textbf{Artifact-pair experience.}
\sys{} represents artifact-pair experiences using \emph{relationships}, which describe structural connections between pairs of artifacts.
For example, relationships between a document and another document or table may be defined by overlapping entities (e.g., ``\texttt{sales2} and \texttt{costs1.pdf} are connected via Delaware'' from~\encirclefigone{C}), while relationships between two tables may be defined by joinable columns.
This design allows \sys{} to capture meaningful co‑occurrences between artifacts so that models can be proactively introduced to the neighbors of relevant artifacts. These are neighboring artifacts that might otherwise be missed or require multiple rounds of retrieval and reasoning to discover.

\subsection{Experience retrieval}
\label{sec:expindex-retrieval}
We can now look at \sys{}'s experience retrieval mechanism (\textsc{RetrieveExperience}$(\mathcal{S}, q, \mathcal{A}\textsubscript{ID})$ in \Cref{alg:expindex}), which identifies relevant experiences and their associated artifacts to guide a model's retrieval and reasoning behavior in a new task.
A natural retrieval strategy is to use lexical or semantic similarity between the input queries and past experiences (e.g., descriptions of new and past tasks).
However, we also observe that models are increasingly leveraging file-system-based search tools~\citep{codeact, openhands} to inspect artifacts when solving tasks.
This behavior creates an opportunity to also search over past experiences about artifacts involved with such tools.
Accordingly, our experience retrieval process takes as input a text query $q$ and, optionally, a set of \texttt{artifact\_id}s $\mathcal{A}\textsubscript{ID}$. To return information that is both \textit{relevant} and \textit{complete}, the retrieval process proceeds as follows. We first retrieve a set of candidate artifacts by consulting single-artifact experience. We then employ an off-the-shelf LLM to assess the relevance of each candidate artifact to $q$. For candidate artifacts deemed relevant, we further consult artifact‑pair experience to introduce their neighboring artifacts.



\textbf{Candidate artifacts.}
Taking the set of all \textbf{single-artifact experiences} $E$, we embed the input text query $q$ using an off-the-shelf embedding model and retrieve the top-$k$ most relevant experiences from $E$ along with their associated artifact.
We compute relevance using a combined similarity between the query embedding and the embeddings of each past experience's \textit{task description} and \textit{contribution summary}.
If a set of \texttt{artifact\_id}s $\mathcal{A}\textsubscript{ID}$ is provided, we additionally perform artifact-conditioned retrieval. For each $a{\textsubscript{ID}} \in \mathcal{A}\textsubscript{ID}$, we consider only the single-artifact experiences associated with $a\textsubscript{ID}$, denoted $E_a$, and select the top-$k$ experiences from $E_a$ based on the same combined similarity between the query embedding and the embeddings of the past task descriptions and contribution summaries.
We then union the artifacts retrieved through $q$ and $\mathcal{A\textsubscript{ID}}$ (if available) to form a set of candidate artifacts.

\textbf{Artifact relevance analysis.}
To ensure that the candidate set only contains \textit{relevant} artifacts and experiences, we apply an off-the-shelf LLM (\texttt{GPT-5-Mini} by default) to classify each candidate artifact, based on its associated experiences, as \texttt{relevant}, \texttt{irrelevant}, or \texttt{inconclusive} for the input query $q$.
If an artifact is deemed \texttt{relevant}, the LLM also identifies which specific experiences are relevant. We then prune the artifacts and experiences deemed \texttt{irrelevant}. We provide detailed prompts in \Cref{app:prompt-analysis}.
As shown in the example in \Cref{fig:diagram}, \texttt{sales1} is determined \texttt{irrelevant} because from experience~\encirclefigone{A}, its time period does not overlap with the required period of 2020-2025, while \texttt{sales2} is determined \texttt{relevant} because from experience~\encirclefigone{B}, its time period overlaps.

\textbf{Neighboring artifacts.}
To improve \textit{completeness}, we consult the \textbf{artifact‑pair experience}, which captures co‑occurrence patterns and \textit{relationships} between artifacts. For each candidate artifact deemed \texttt{relevant}, we fetch its neighboring artifacts and apply the relevance analysis described above. We retain only the neighbors classified as \texttt{relevant}.
For instance, \texttt{costs1.pdf} is retrieved as a neighboring artifact of \texttt{sales2} via the connecting entity ``Delaware'' shown in~\encirclefigone{C}.

\subsection{Experience injection}
\label{sec:expindex-injection}
Once relevant artifacts and associated experiences are retrieved, they are used to guide the model's behavior.
Under a typical agentic setting shown in \Cref{alg:expindex} with a predefined search framework, we identify two primary non-parametric components that influence model behavior and remain mutable at runtime: (i) the outputs of search frameworks, and (ii) the model prompt context.
However, both components, by default, operate without awareness of prior experiences. To bridge this gap, \sys{} provides a unified middleware layer that injects retrieved experiences (\Cref{sec:expindex-retrieval}) by enriching both searched artifacts and the model prompt before they reach the model.
Because of the lightweight and non‑invasive nature of middleware, \sys{} can be easily integrated with common search frameworks, such as retrieve-then-rerank pipelines in traditional retrieval‑augmented generation systems~\citep{lewis2020retrieval}, retrieval tools invoked within agentic systems~\citep{yao2022react}, and agentic search with file system tools~\citep{openhands}.

\textbf{Search framework output.}
A standard search framework takes a query $q$ as input and returns a set of artifacts $\mathcal{A}$ together with their IDs $\mathcal{A}\textsubscript{ID}$ (line 9 in \Cref{alg:expindex}) to the model.
Our middleware enriches this output by intercepting $\mathcal{A}$ before it reaches the model: it performs experience retrieval using $q$ and $\mathcal{A}\textsubscript{ID}$, and then modifies the tool output based on the retrieved experience.
In the example shown in \Cref{fig:diagram}, for task 2 without \sys{}, the search tool returns $\mathcal{A}$, which includes \texttt{sales1} and \texttt{sales15}, introducing irrelevant artifacts and incomplete information. With \sys{}, $\mathcal{A}$ is refined: \texttt{sales1} is pruned, and \texttt{sales2} and \texttt{costs1.pdf} are introduced before passed to the model.
We give the details of our tool-response modification mechanism in \Cref{app:expindex-injection}.

\textbf{Model prompt.}
Our middleware also enriches the model prompt to complement tool response enrichment. Given the task $t$, we perform experience retrieval based on $t$
and inject the relevant experiences directly into the system prompt using the formatting described in \Cref{app:prompt-system}. Placing relevant experiences in the initial system prompt exposes the agent to useful prior knowledge before it performs extensive reasoning and issues tool calls, enabling earlier guidance and reducing the risk of suboptimal reasoning caused by delayed access to helpful experiences.\\
\textit{Lightweight reflection.}
The prompt also includes a lightweight reflection step that instructs the agent to (1) assess the relevance of the retrieved experience and artifacts and disregard irrelevant ones, and (2) verify experiences against the current artifacts, giving them priority over past experiences in the event of discrepancies caused by hallucinations, errors, or outdated information.





%% file: sections/exp.tex
\section{Experiments}
\label{sec:experiments}

\subsection{Setup}\label{sec:experimental-setup}

%

\begin{table}[!htb]
\centering
\scriptsize
\caption{Details on datasets and baseline solutions.}
\vspace{0.5em}
\begin{tabularx}{\textwidth}{@{} l p{0.6in} l l l X @{}}
\toprule
\textbf{Dataset} & \textbf{Domain} & \textbf{Modality} & \textbf{Task type} & \textbf{Baseline solution} & \textbf{Search frameworks}\\
\midrule
\makecell[l]{Locomo\\~\citep{maharana2024evaluating}} & General & Files & Conversational QA & \makecell[l]{ReAct\\~\citep{yao2022react}} & File-system search (e.g., \texttt{bash})\\
SWE-bench-ver.\\~\citep{jimenez2023swe} & Tech & Files & Bug fixes & ReAct & File-system search\\
\midrule
\makecell[l]{SQA\\~\citep{bragg2025astabench}} & Science & Documents & Long-form generation & \makecell[l]{ScholarQA\\~\citep{singh2025ai2}} & Custom retrieve-and-rerank to\newline Semantic Scholar\\
\makecell[l]{Freshstack\\~\citep{thakur2025freshstack}} & Technical doc. & Documents & Long-form generation & ReAct & Vector search tool\newline(\texttt{Qwen3-0.6B-Embedding})\\
\makecell[l]{Musique\\~\citep{trivedi2022musique}} & General & Documents & Multi-hop QA & ReAct & Vector search tool\newline(\texttt{Qwen3-8B-Embedding})\\
\midrule
\makecell[l]{Spider\\~\citep{yu2018spider}} & Misc. & DB Tables & Text-to-SQL & ReAct & Vector search tool (\texttt{Q8B-Embed}),\newline SQL execution tool\\
\makecell[l]{MMQA\\~\citep{wu2025mmqa}} & Misc. & DB Tables & Factoid QA & ReAct & Vector search tool (\texttt{Q8B-Embed}),\newline SQL execution tool\\
\bottomrule
\end{tabularx}
\label{tab:dataset}
\end{table}

\textbf{Datasets and baseline solutions.}
As discussed in \Cref{sec:intro}, \sys{} is designed with two forms of generality: it supports a broad spectrum of workloads that operate over a \textit{shared} corpus of artifacts, and it integrates with diverse search frameworks used across different solutions. To evaluate this generality, we selected a suite of knowledge‑intensive tasks spanning multiple domains, modalities, and task types.
As the baseline solution, we use the state‑of‑the‑art ScholarQA~\citep{singh2025ai2} for the SQA dataset, and ReAct~\citep{yao2022react} for all remaining datasets, as it represents the standard agentic approach. Our ReAct implementation is augmented with the commonly adopted set of search tools described in~\cite{chang2026karl}.
\Cref{tab:dataset} shows a summary of these datasets and baselines (details in \Cref{app:eval}).


\textbf{Baselines.}
We compare against two baselines, one from each category of related work discussed in~\Cref{sec:related}: long-term memory and offline enrichment. Like \sys{}, both generalize across a wide range of datasets.
\begin{itemize}
    \item \textit{Memory baseline.} We adopt \reasoningbank{}~\citep{ouyang2025reasoningbank} as a representative abstracted memory baseline, which extracts reasoning patterns from previous tasks and reuses them on new ones. Given past reasoning traces, \reasoningbank{} extracts generalizable insights into discrete \textit{memory items} and stores them in a memory store. When a new task arrives, it retrieves the memory items linked to semantically similar prior tasks and appends them to the system prompt to guide the model's reasoning.
    \item \textit{Offline enrichment baseline.} We adopt \enrichindex{}~\citep{chen2025enrichindex} as a representative offline enrichment approach. In an offline phase, it uses an LLM to enrich each artifact in the corpus (e.g., a document, table, or code file) with three additional labels: a content summary, a description of the artifact's purpose, and a set of potential QA pairs. During online task-solving, artifacts are retrieved using both their original content and these labels, and are then provided to the agent in the same manner as in \sys{} (\Cref{sec:expindex-injection}). As noted in \Cref{sec:related}, this approach requires access to the entire corpus in advance, which is impractical for SQA, where queries are issued online against Semantic Scholar's corpus of over 238 million papers. We therefore do not report \enrichindex{} results on SQA.
\end{itemize}

\textbf{Evaluation with and without memory.}
For each memory solution (\reasoningbank{} and \sys{}), we build a memory store by splitting the dataset into \textit{experience tasks} and \textit{evaluation tasks}. By default, experience tasks contain 80\% of the dataset (we also vary this proportion in \Cref{fig:task-pct-sensitivity}) and evaluation tasks include the rest. 
We first run the baseline solution on all experience tasks (model used here is denoted as the \textit{experience model}), and collect generated reasoning traces. Each memory solution then extracts reusable experience and insights from these traces and indexes them into its memory store.
We then evaluate the baseline solution on the evaluation tasks (model used here is denoted as the \textit{evaluation model}), both with and without memory.
We use a static memory store instead of updating it continuously (e.g., after every $N$ tasks) to enable controlled comparisons, particularly to study how performance depends on the number of stored experiences and on the experience model (\Cref{sec:analysis}).

\textbf{Metrics.}
We measure performance and resource usage in two phases. The \textit{online phase} covers solving the evaluation tasks, and the \textit{indexing phase} covers building the additional indexes each method uses, i.e., constructing the memory store from experience-task traces (\reasoningbank{} and \sys{}) or enriching the corpus offline (\enrichindex{}). All dollar costs are computed by tracking input and output token usage and applying the corresponding API pricing.
\begin{itemize}
    \item \textit{Online metrics.} For the online phase, we report the average \textit{answer score} and the average \textit{online cost}. The answer score is computed by comparing the predicted answer to the ground-truth answer (details in~\Cref{app:eval}). The online cost includes all runtime expenses incurred while solving the evaluation tasks, including any charges for retrieving from the memory store or enriched index, and is reported in dollars per \textit{thousand} tasks.
    \item \textit{Indexing metrics.} For the indexing phase, we report the \textit{indexing cost} and the \textit{storage footprint}. The indexing cost is the expense of extracting information from experience-task traces and populating the memory store, or of enriching the corpus offline (details in~\Cref{app:expindex-indexing}). The storage footprint is the size of the additional indexes each method maintains beyond the original corpus. The baseline solution builds no additional indexes, so it incurs no indexing cost and has no storage footprint.
\end{itemize}

\subsection{Results}

\begin{table}
\centering
\caption{Performance on files, codebases, documents, and tables. We report cost in dollars per thousand tasks.
To compute cost, we track the number of input and output tokens and apply the appropriate API pricing.
\texttt{Base} stands for the baseline solutions, \texttt{Base+X} stands for augmenting the baseline solutions with \texttt{X}. \texttt{RB} stands for \reasoningbank{}~\citep{ouyang2025reasoningbank}. \texttt{EI} stands for \enrichindex{}~\citep{chen2025enrichindex}. \texttt{EXP} stands for our method, \sys{}.
Higher scores are better and lower costs are better.
$\dagger$\enrichindex{} is not evaluated on SQA, since it requires offline access to the full corpus, whereas SQA queries Semantic Scholar's 238M+ papers online.
*Cross-task generalization between text-to-SQL tasks and factoid QA over a shared database. MMQA is evaluated using a memory store constructed from the reasoning traces of Spider tasks.
}
\vspace{0.5em}
\scriptsize
\begingroup
\setlength{\tabcolsep}{3pt}
\begin{NiceTabularX}{\textwidth}{@{}Xrr rr rr rr rr rr rr@{}}
& \multicolumn{2}{c}{\textbf{Locomo}} & \multicolumn{2}{c}{\textbf{SWE-bench-ver.}} & \multicolumn{2}{c}{\textbf{SQA}} & \multicolumn{2}{c}{\textbf{Freshstack}} & \multicolumn{2}{c@{}}{\textbf{Musique}} & \multicolumn{2}{c@{}}{\textbf{Spider (Text-to-SQL)}} & \multicolumn{2}{c@{}}{\textbf{MMQA* (Factoid QA)}}
\\
\cmidrule(lr){2-3} \cmidrule(lr){4-5} \cmidrule(lr){6-7} \cmidrule(lr){8-9} \cmidrule(lr){10-11} \cmidrule(lr){12-13} \cmidrule(l){14-15}
& Score & Cost (\$) & Score & Cost & Score & Cost & Score & Cost & Score & Cost & Score & Cost & Score & Cost \\
\midrule
\multicolumn{15}{@{}l}{
\rowcolor{gray!10}
\texttt{Gemini-3-Flash} (experience and evaluation model)} \\
\midrule
\texttt{Base} & 88.8 & 155.2 & 76.5 & 990.9 & 88.7 & 21.6 & 73.0 & 40.1 & 67.0 & 91.2 & 71.0 & 37.1 & 91.0 & 61.3 \\
\texttt{Base+RB} & 87.9 & 125.6 & 76.5 & 883.3 & 87.1 & 22.5 & 70.1 & 38.3 & 74.0 & 115.4 & 70.0 & 41.4 & 92.0 & 54.8 \\
\texttt{Base+EI} & 86.0 & 105.3 & 78.6 & 919.8 & ---\dag & ---\dag & 74.0 & 42.9 & 70.0 & 40.9 & 72.0 & 35.6 & 90.0 & 42.6 \\
\texttt{Base+EXP} & \textbf{92.5} & \textbf{93.4} & \textbf{81.6} & \textbf{782.0} & \textbf{90.6} & \textbf{21.0} & 73.3 & \textbf{33.8} & \textbf{78.0} & 66.5 & \textbf{75.0} & \textbf{22.0} & \textbf{92.0} & \textbf{32.3} \\
\midrule
\multicolumn{15}{@{}l}{
\rowcolor{gray!10}
\texttt{GPT-5-Mini} (experience and evaluation model)} \\
\midrule
\texttt{Base} & 82.2 & 14.2 & 58.2 & 27.9 & 81.8 & 45.0 & 66.6 & 23.6 & 62.0 & 8.45 & 57.0 & 2.27 & 84.0 & 3.47 \\
\texttt{Base+RB} & 83.2 & 13.4 & 59.2 & 29.6 & 80.0 & 58.3 & 67.5 & 23.6 & 67.0 & 9.54 & 57.0 & 3.86 & 85.0 & 4.80 \\
\texttt{Base+EI} & 88.8 & 7.66 & 61.2 & 26.9 & ---\dag & ---\dag & 67.9 & 19.2 & 63.0 & 7.70 & 60.0 & 1.91 & 87.0 & 2.96 \\
\texttt{Base+EXP} & \textbf{88.8} & \textbf{7.03} & 60.2 & \textbf{22.1} & \textbf{85.0} & 45.5 & \textbf{70.2} & 22.1 & \textbf{68.0} & 8.91 & \textbf{65.0} & 2.88 & \textbf{89.0} & 3.69 \\
\bottomrule
\end{NiceTabularX}
\endgroup
\label{tab:perf-1}
\vspace{-1em}
\end{table}

\Cref{tab:perf-1} presents the answer scores and online costs of all methods across all datasets and models.
Two key findings emerge:
(i) \textbf{\sys{} achieves the highest average score at the lowest average online cost.}
Averaged across both evaluation models, \sys{} attains the highest answer score, a relative improvement of 3.9\% over the strongest baseline, \enrichindex{}. It also has the lowest online cost, reducing cost by 20.8\% relative to the baseline solutions. These results underscore the benefits of artifact-grounded memory over no memory, abstracted memory, and offline enrichment.
(ii) \textbf{Results on Spider and MMQA demonstrate cross‑task generalization.}
MMQA (a factoid QA dataset) and Spider (a text‑to‑SQL dataset) share the same table corpus, and the MMQA memory store is built entirely from Spider’s reasoning traces. The fact that \sys{} can leverage this memory to improve accuracy and reduce cost shows that experiences grounded in a shared corpus transfer effectively across task types.

Furthermore, \Cref{tab:index-cost} shows that \textbf{\sys{} reduces average indexing cost by $115.0\times$ and storage footprint by $73.0\times$ relative to the strongest baseline, \enrichindex{}.}
This is because \enrichindex{} requires scanning every document, causing both financial cost and storage footprint to scale directly with corpus size. This makes it inefficient for large corpora or scenarios where full-corpus access is impractical, as with the live Semantic Scholar corpus in SQA. In contrast, \sys{}'s costs scale primarily with the number of tasks for which experiences are stored, not with corpus size.
Compared with \reasoningbank{}, \sys{} has a comparable indexing cost and a storage footprint of the same order of magnitude, while \Cref{tab:perf-1} shows that it improves the average answer score by 5.01\% and reduces average online cost by 19.8\%.

\begin{wraptable}{l}{0.5\textwidth}
\vspace{-2em}
\centering
\caption{Indexing cost (\$) and storage footprint (MB) of each method, averaged across datasets.}
\vspace{-0.5em}
\scriptsize
\begingroup
\setlength{\tabcolsep}{3pt}
\begin{NiceTabularX}{\linewidth}{@{}Xrr rr rr@{}}
& \multicolumn{2}{c}{\textbf{Base+RB}} & \multicolumn{2}{c@{}}{\textbf{Base+EI}} & \multicolumn{2}{c@{}}{\textbf{Base+EXP}} \\
\cmidrule(lr){2-3} \cmidrule(l){4-5} \cmidrule(l){6-7}
& Cost & Storage & Cost & Storage & Cost & Storage \\
\midrule
\texttt{Gemini-3-Flash} & 0.503 & 4.20 & 93.4 & 1108.7 & 0.628 & 15.5 \\
\texttt{GPT-5-Mini} & 0.398 & 4.19 & 34.2 & 1110.7 & 0.421 & 14.9 \\
\bottomrule
\end{NiceTabularX}
\endgroup
\label{tab:index-cost}
\vspace{-2em}
\end{wraptable}

\paragraph{Qualitative examples.}
We manually reviewed the agents' task-solving traces and selected representative examples illustrating how \sys{} benefits the base agent through both single-artifact and artifact-pair experiences (\Cref{sec:expindex-representation}). Details are provided in~\Cref{app:qualitative}.

%% file: sections/analysis.tex
\section{Analysis}
\label{sec:analysis}




\begin{figure}[t]
  \begin{minipage}{0.49\textwidth}
    \scriptsize
    \centering
    \begin{tabularx}{0.95\textwidth}{@{}Xllrr@{}}
     & \makecell[l]{\textbf{Experience}\\\textbf{model}} & \makecell[l]{\textbf{Evaluation}\\\textbf{model}} & \makecell[r]{\textbf{Score}\\\ } & \makecell[r]{\textbf{Online}\\\textbf{Cost (\$)}} \\
    \midrule
    \makecell[l]{\texttt{Base}\\\ } & \makecell[l]{---\\\ } & \makecell[l]{\texttt{GPT-5-}\\\texttt{Mini}} & \makecell[r]{82.2\\\ } & \makecell[r]{14.2\\\ }\\
    \makecell[l]{\texttt{Base}\\\ } & \makecell[l]{---\\\ } & \makecell[l]{\texttt{Gemini-3-}\\\texttt{Flash}} & \makecell[r]{88.8\\\ } & \makecell[r]{155.18\\\ }\\
    \makecell[l]{\texttt{Base}+\texttt{EXP}\\\ } & \makecell[l]{\texttt{GPT-5-}\\\texttt{Mini}} & \makecell[l]{\texttt{GPT-5-}\\\texttt{Mini}} & \makecell[r]{88.8\\\ } & \makecell[r]{7.03\\\ }\\
    \makecell[l]{\texttt{Base}+\texttt{EXP}\\\ } & \makecell[l]{\texttt{Gemini-3-}\\\texttt{Flash}} & \makecell[l]{\texttt{Gemini-3-}\\\texttt{Flash}} & \makecell[r]{\textbf{92.5}\\\ } & \makecell[r]{93.4\\\ } \\
    \makecell[l]{\texttt{Base}+\texttt{EXP}\\\ } & \makecell[l]{\texttt{Gemini-3-}\\\texttt{Flash}} & \makecell[l]{\texttt{GPT-5-}\\\texttt{Mini}} & \makecell[r]{\textbf{92.5}\\\ } & \makecell[r]{7.50\\\ }\\
    \bottomrule
    \end{tabularx}
    \captionof{table}{Experience as teacher-student learning. Experiences from a stronger model (\texttt{Gemini-3-Flash}) enable a weaker model (\texttt{GPT-5-Mini}) to reach comparable performance.}\label{tab:perf-3}
  \end{minipage}%
  \hfill
  \begin{minipage}{0.48\textwidth}
    \centering
    \includegraphics[width=\textwidth]{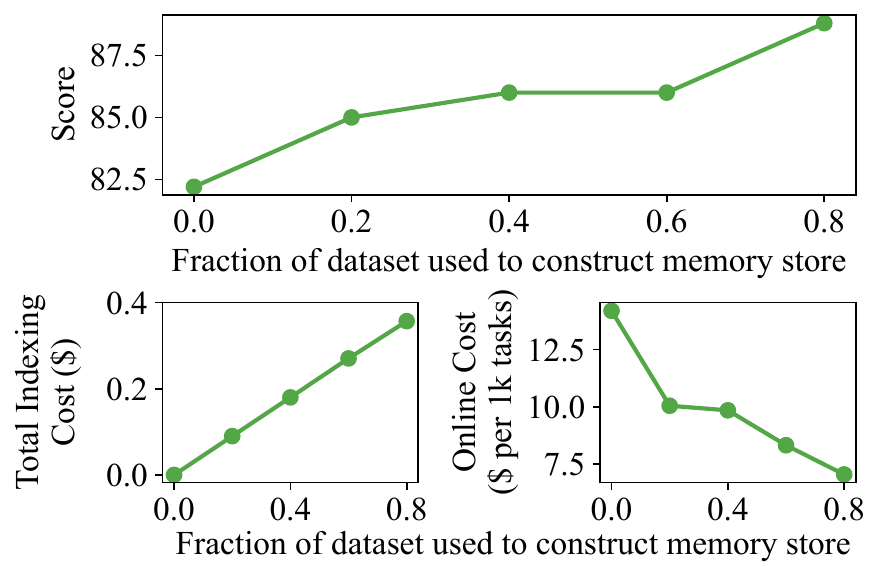}
    \vspace{-1em}
    \caption{\sys{} improves accuracy and decreases online cost as more tasks are indexed.}
    \label{fig:task-pct-sensitivity}
  \end{minipage}
  \vspace{-1em}
\end{figure}


\Cref{sec:experiments} demonstrates that \sys{} outperforms the baseline solutions, abstracted memory, and offline enrichment. This advantage comes from \sys{}'s ability to reuse artifact knowledge captured from prior tasks. Such prior experience depends on two factors:
(i) \textit{the experience model} that originally produced the reasoning traces that are later extracted and indexed as experiences, and
(ii) the number of experiences in the memory store.
In this section, we analyze \sys{}'s performance along both dimensions using the Locomo dataset.
We also analyze sensitivity to the model used for artifact relevance analysis (\Cref{sec:expindex-retrieval}), and perform an ablation that removes the lightweight reflection step (\Cref{sec:expindex-injection}).

\textbf{Experience as teacher-student learning.}
In \Cref{sec:experiments}, the experience model used to solve the experience tasks is the same as the evaluation model used to solve the evaluation tasks.
\Cref{tab:perf-3} examines a different setting: a stronger experience model (\texttt{Gemini-3-Flash}) paired with a weaker evaluation model (\texttt{GPT-5-Mini}).
We observe that, for the same evaluation model, performance improves as the experience model improves. Remarkably, the weaker evaluation model paired with the stronger experience model reaches performance comparable to the stronger evaluation model, but at substantially lower cost. This result shows that experiences serve as a mechanism for teacher-student learning, enabling a weaker model to benefit from experiences of a stronger model.

\textbf{Performance varying size of memory store.}
Experiments in \Cref{sec:experiments} used 80\% of the dataset to construct the memory store $\mathcal{S}$. To test the robustness of \sys{} to different sizes of $\mathcal{S}$, we vary the proportion of the dataset to 20\%, 40\%, and 60\%. \Cref{fig:task-pct-sensitivity} summarizes the results. Two patterns emerge:
(i) Performance at all sizes of $\mathcal{S}$ outperforms the 0\% setting (i.e., the baseline solution without memory), indicating that \sys{} is robust and consistently benefits from experience.
(ii) As the size of $\mathcal{S}$ increases, both answer quality and cost efficiency improve, showing that \sys{} effectively leverages larger experience pools to deliver better outcomes.



\begin{wraptable}{l}{0.5\textwidth}
\centering
\caption{Score using different relevance models.}
\scriptsize
\begingroup
\setlength{\tabcolsep}{3pt}
\begin{NiceTabularX}{\linewidth}{@{}Xrrr@{}}
& \textbf{GPT-5-Mini} & \textbf{Nemotron-3-Ultra} & \textbf{Mimo-V2.5} \\
\midrule
\texttt{Gemini-3-Flash} & 92.5 & 92.5 & 91.6\\
\texttt{GPT-5-Mini} & 88.8 & 89.7 & 88.8 \\
\bottomrule
\end{NiceTabularX}
\endgroup
\label{tab:perf-relevance-model}
\vspace{-1em}
\end{wraptable}

\textbf{Sensitivity to the artifact-relevance model.}
As mentioned in~\Cref{sec:expindex-retrieval}, we utilized \texttt{GPT-5-Mini} for artifact relevance analysis and presented its results in \Cref{sec:experiments}. To demonstrate robustness, we evaluated two highly popular open-source alternatives based on usage rankings from OpenRouter: \texttt{Nemotron-3-Ultra}~\citep{blakeman2026nemotron} and \texttt{Mimo-V2.5}~\citep{team2026full}. \Cref{tab:perf-relevance-model} shows that both alternative models achieve performance comparable to \texttt{GPT-5-Mini}.

\begin{wraptable}{l}{0.5\textwidth}
\centering
\caption{Score with and without reflection.}
\scriptsize
\begingroup
\setlength{\tabcolsep}{3pt}
\begin{NiceTabularX}{\linewidth}{@{}Xrr@{}}
& \textbf{With reflection} & \textbf{Without reflection} \\
\midrule
\texttt{Gemini-3-Flash} & \textbf{92.5} & 91.6\\
\texttt{GPT-5-Mini} & \textbf{88.8} & 85.0\\
\bottomrule
\end{NiceTabularX}
\endgroup
\label{tab:perf-reflection}
\vspace{-1em}
\end{wraptable}

\textbf{Ablation of lightweight reflection.}
As mentioned in \Cref{sec:expindex-injection}, \sys{} uses a lightweight reflection step that prompts the task-solving model to (1) assess the relevance of the retrieved experience and artifacts and disregard irrelevant ones, and (2) verify experiences against the current artifacts, giving them priority over past experiences in the event of discrepancies caused by hallucinations, errors, or outdated information. \Cref{tab:perf-reflection} shows that removing this reflection step leads to a drop in answer score, indicating that reflection helps mitigate noise and errors.








%% file: sections/conclusion.tex
\section{Conclusion}

Memory systems aim to give agentic AI the statefulness needed to learn from prior tasks. Although existing work explores many forms of ``memory,'' such as user preferences, factual attributes, and general reasoning patterns, these approaches fall short for a common and highly impactful scenario: knowledge‑intensive tasks over a shared corpus of artifacts, where humans naturally accumulate experiential knowledge about those artifacts through repeated task solving.
\sys{} addresses this gap by introducing \textit{artifact‑grounded memory}, a framework that captures and reuses knowledge about artifacts derived from past reasoning traces.
It represents experience at two levels: single‑artifact experience, which records how an artifact contributed to earlier tasks, and artifact‑pair experience, which captures structural relationships between artifacts. When a new task arrives, \sys{} retrieves these experiences, aiming to rapidly surface the full set of relevant artifacts.
Across a wide range of tasks and corpora, including scientific literature, conversational histories, and codebases, our method consistently improves answer quality while reducing computational cost. We further observe that artifact‑grounded experience supports cross‑task generalization and enables effective teacher-student learning, underscoring its broader utility. We hope this work encourages a shift from non‑grounded memory designs toward memory representations that are explicitly tied to artifacts.

%% file: sections/appendix.tex
\section{Implementation details of \sys{}}
\label{app:expindex-implementation}

\subsection{Experience injection}
\label{app:expindex-injection}
As described in \Cref{sec:expindex-injection}, a standard search framework takes a query $q$ as input and returns a set of artifacts $\mathcal{A}$ along with their IDs $\mathcal{A}\textsubscript{ID}$.
Using $q$ and $\mathcal{A}\textsubscript{ID}$, we perform experience retrieval to obtain artifacts $\mathcal{A}\textsubscript{exp}$ with associated past experiences signals. For each \texttt{artifact\_id}, these signals include its classification as \texttt{relevant}, \texttt{irrelevant}, or \texttt{inconclusive}, as well as associated experience snippets (\Cref{sec:expindex-representation}).
We then modify the original set of artifacts $\mathcal{A}$ to become $\mathcal{A'}$ as follows:
\begin{itemize}
\item Artifacts in $\mathcal{A}\textsubscript{exp}$ classified as \texttt{irrelevant} are removed if they were in $\mathcal{A}$
\item Artifacts classified as \texttt{relevant} are either retained (if already in $\mathcal{A}$) or introduced (if not originally in $\mathcal{A}$), and their content enriched with relevant experience snippets
\item Artifacts classified as \texttt{inconclusive} remain unchanged if they were in $\mathcal{A}$
\end{itemize}
The model then receives $\mathcal{A'}$ instead of the original $\mathcal{A}$.

\subsection{Experience indexing}
\label{app:expindex-indexing}

Here, we look at how \sys{} extracts experiences from past reasoning traces and indexes them into a persistent memory store, which enables it to accumulate experience over time.

\textbf{Extraction from reasoning traces.}
We extract both single‑artifact and artifact‑pair experience from the reasoning traces generated during past task solving. Concretely, we construct a complete reasoning trace by concatenating all rounds of the model's internal reasoning~\citep{jaech2024openai}, explicit thinking traces~\citep{yao2022react}, and final answers, including any citations to artifacts~\citep{singh2025ai2}.
From this full trace, we process each mentioned artifact by prompting an off-the-shelf LLM (\texttt{GPT-5-Mini} by default) to generate its \textit{contribution summary} and to extract \textit{snippets} based on the cited content. Using the set of all artifacts referenced in the trace, we then prompt the same LLM to infer \textit{relationships} between artifact pairs.
We provide full prompt details in \Cref{app:prompt-extraction}.
In the running example in \Cref{fig:diagram}, the reasoning trace for task 1 mentions artifacts \texttt{sales1}, \texttt{sales2}, \texttt{costs1.pdf}. An LLM then extracts from the trace single-artifact experiences such as~\encirclefigone{A} and~\encirclefigone{B}, as well as artifact-pair experiences such as~\encirclefigone{C}.

\textbf{Indexing.}
We store single-artifact and artifact-pair experiences in a database (DuckDB~\citep{raasveldt2019duckdb} in our case). 
After a batch of tasks completes, we apply the extraction process described above to the reasoning traces of every completed task, producing a set of single‑artifact and artifact‑pair experiences.
To support retrieval (\Cref{sec:expindex-retrieval}), we compute embeddings of the \textit{task descriptions} and \textit{contribution summaries} using an off-the-shelf text embedding model (\texttt{Qwen3-8B-Embedding}~\citep{zhang2025qwen3}) and then insert these experiences and embeddings into the database.

\clearpage

\section{Prompts}
\label{app:prompt}

\subsection{Prompt for artifact relevance analysis}
\label{app:prompt-analysis}

We provide the prompt we use to analyze artifact relevance in \Cref{tab:prompt-analysis}.

\begin{table}[h!]
\caption{Prompt for analyzing artifact relevance.}
\label{tab:prompt-analysis}

\begin{tcolorbox}[title=Artifact relevance analysis]
Given a question and a set of artifacts, each including a list of prior experiences (where each experience consists of a contribution summary, and associated snippets from the artifact), your task is to evaluate, for each artifact, whether its prior experience is (1) relevant to the provided question; (2) irrelevant to the provided question; or (3) inconclusive.\\
\\
For each artifact, you should also evaluate its relevance based on its relationships with previously identified useful artifacts: <\texttt{artifact\_id}s>\\
\\
Your output should be a JSON list containing, for each artifact, the artifact ID, your judgment ("relevant", "irrelevant", or "inconclusive"), and, when applicable, the IDs of the relevant experience.\\
\\
Question:\\
<question>\\
\\
Artifacts:\\
<artifacts>
\end{tcolorbox}

\end{table}

\clearpage

\subsection{Prompt for injecting experiences in model system prompt}
\label{app:prompt-system}

We provide the system prompt we use when injecting retrieved experiences in \Cref{tab:prompt-system}.

\begin{table}[h!]
\caption{System prompt for injecting retrieved experiences.}
\label{tab:prompt-system}

\begin{tcolorbox}[title=System message]
You are given a set of artifacts with past experiences, each containing an artifact ID, past user questions, descriptions of how the artifact was used to answer them, and the specific snippets that were used.\\
\\
When handling a new task, you should:\\
(1) Consult the experiences as prior examples. Use them to guide your reasoning, retrieval strategy, and expectations about how artifacts may be relevant.\\
(2) Determine relevance. Identify which, if any, of the past experiences apply to the new task. If none are relevant, continue normally without forcing a connection.\\
(3) Always verify against the current artifacts. The experiences are helpful context, but the actual artifacts available now take priority. If there is any discrepancy, updated or current file contents override the past experiences.\\
\\
Past experiences:\\
<artifact id=`...'>\\
<experience>\\
Question: ...\\
\\
Contribution summary: ...\\
\\
Snippets: ...\\
</experience>\\

<experience>\\
...\\
</experience>\\
</artifact>
\end{tcolorbox}
\end{table}

\clearpage

\subsection{Prompt for extracting experiences from reasoning traces}
\label{app:prompt-extraction}

We provide the prompt we use to generate contribution summaries and relationships in \Cref{tab:prompt-extraction}.

\begin{table}[h!]
\caption{Prompt for generating contribution summary and relationships.}
\label{tab:prompt-extraction}

\begin{tcolorbox}[title=Single-artifact experience]
Given a question and its corresponding response, summarize how each artifact in <artifact IDs> contributes to addressing the question. For each artifact, include its artifact\_id and a concise description of its contribution in a few sentences.\\
\\
Question:\\
<question>\\
\\
Response:\\
<response>
\end{tcolorbox}

\begin{tcolorbox}[title=Artifact-pair experience]
Given a question and its corresponding response, identify direct relationships between pairs of artifacts mentioned in the response. Do not include pairs of artifacts with no direct relationships.\\
\\
Question:\\
<question>\\
\\
Response:\\
<response>
\end{tcolorbox}

\end{table}

\clearpage

\section{Qualitative examples}
\label{app:qualitative}

\subsection{Overview}
To show how \sys{} works and the benefits of our experience representation in \Cref{sec:expindex-representation}, we conduct a brief qualitative analysis with concrete examples.

\begin{figure}[h]
  \vspace{-0.75em}
  \centering
  \begin{overpic}[width=\textwidth, clip, trim={0cm 1.9in 0cm 0cm}]{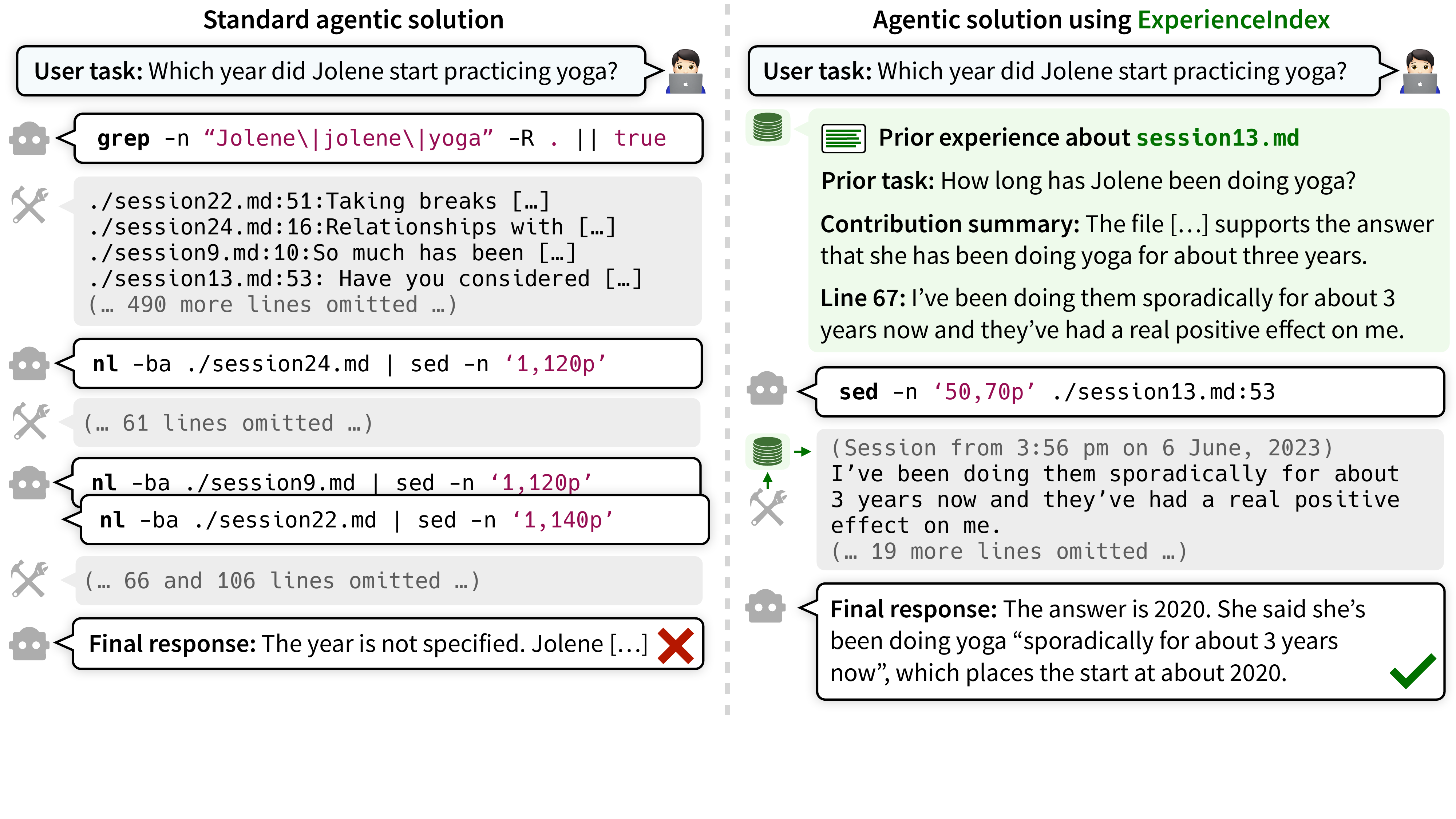}
    \put(45,34.5){\captionlabelsmall{A}} 
    \put(45,23.5){\captionlabelsmall{B}} 
    \put(45.25,1){\captionlabelsmall{C}} 
    \put(96,39){\captionlabelsmall{D}} 
    \put(96,21.6){\captionlabelsmall{E}} 
    \put(96,6.5){\captionlabelsmall{F}} 
  \end{overpic}
  \vspace{-0.75em}
  \caption{An example Locomo task where \sys{} surfaces files and snippets from relevant prior single-artifact experiences, which helps the agent efficiently reach the correct answer.}
  \label{fig:single-artifact-locomo}
\end{figure}

\begin{figure}[h]
  \centering
  \begin{overpic}[width=\textwidth, clip, trim={0cm 3.5in 0cm 0cm}]{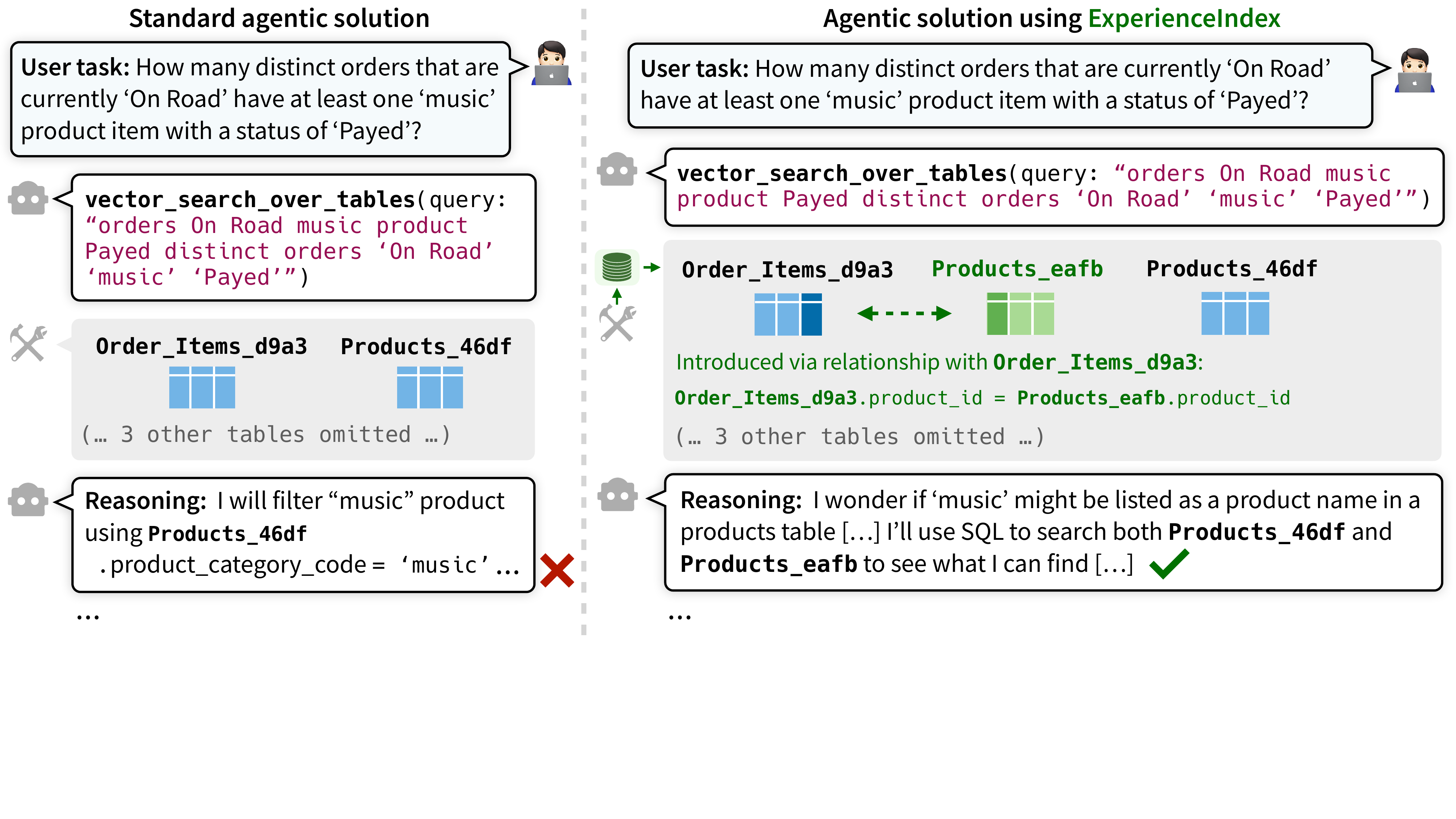}
    \put(37,19){\captionlabelsmall{A}} 
    \put(37,8){\captionlabelsmall{B}} 
    \put(74,21){\captionlabelsmall{C}} 
    \put(96,4){\captionlabelsmall{D}} 
  \end{overpic}
  \vspace{-0.75em}
  \caption{An example MMQA task where \sys{} surfaces a table the vector search missed (\texttt{Products\_eafb}), which ultimately helps the agent reach the correct answer.}
  \label{fig:pair-mmqa}
  \vspace{-0.5em}
\end{figure}

\textbf{Single-artifact experience filters irrelevant information.}
\Cref{fig:single-artifact-locomo} presents a Locomo example illustrating how current agentic solutions become incorrect and inefficient when handling long conversation histories (stored as files) without \sys{}.
Without knowing which files matter, the model resorts to a generic keyword search that returns both the relevant files and many superficially matching but irrelevant ones, driving up computational cost. Overwhelmed by this noise~\encircle{A}, the model inspects only a subset of relevant files~\encircle{B}, ultimately failing to find the correct answer~\encircle{C}.
In contrast, \sys{} surfaces only the specific files and snippets that matter based on prior experience~\encircle{D}. The model simply checks whether these files contain the necessary information~\encircle{E}, and then produces the answer~\encircle{F}. Here, leveraging single-artifact experience improves answer quality by reducing context load and making task-solving more efficient. The full example appears in \Cref{tab:locomo-react,tab:locomo-expindex}.






\textbf{Artifact-pair experience improves information coverage.}
\Cref{fig:pair-mmqa} shows an example from MMQA, showing how factoid QA over structured tables can fail when prior experience is not reused. Without knowing which tables matter, the model relies on a retrieval tool that returns only a subset of the relevant tables~\encircle{A}. Believing this partial set to be complete, the model proceeds with incomplete evidence and produces an incorrect answer~\encircle{B}. With \sys{}, even when the retriever alone cannot gather all necessary tables, \sys{} introduces neighbor artifacts to surface the full set of relevant tables~\encircle{C}, enabling the model to answer correctly~\encircle{D}. \Cref{tab:mmqa} shows the detailed example.


\subsection{Single-artifact experience}

\Cref{tab:locomo-react} shows a standard agentic solution on a Locomo task. \Cref{tab:locomo-expindex} shows the solution when \sys{} is used on the same task.

\begin{table}[h!]
\caption{\label{tab:locomo-react}Standard agentic soluton on an example Locomo task.}


\begin{tcolorbox}[title=Task]
Which year did Jolene start practicing yoga?
\end{tcolorbox}
\begin{tcolorbox}[title=Assistant messages]
...\\
\textbf{Tool call:} \texttt{grep -n "Jolene\textbackslash{}|jolene\textbackslash{}|yoga" -R /files || true}\\
\textbf{Tool response:}\\
/files/session22\_5:33 pm on 26 August, 2023.md:51:Taking breaks, doing some stretching/yoga, or just going for a walk is really helpful. And don't forget to get enough sleep and take time for self-care. Finding a balance between work and taking care of yourself is important. What self-care activities have you been doing lately?\\
...\\
/files/session24\_2:14 pm on 3 September, 2023.md:16:Relationships with family and friends are so vital. My yoga pals have been my second family - we've held each other up through a lot. The other day I found this old photo. That was when I first started doing yoga. My mum was my biggest fan and source of motivation. She'd often come to my classes with me.\\
...\\
/files/session9\_11:22 am on 13 March, 2023.md:10:So much has been going on lately. I started this yoga class in the neighborhood - it's such a good feeling! Now I get to share the exercise with my neighbors and watch it really transform them.\\
...\\
/files/session13\_3:56 pm on 6 June, 2023.md:53: Have you considered taking some breaks and finding activities like yoga to help you relax and unwind? That might make a difference.\\
(...490 lines omitted ...)\\
\\
\textbf{Tool call:}
\texttt{nl -ba /files/session24\_2:14\textbackslash{} pm\textbackslash{} on\textbackslash{} 3\textbackslash{} September,\textbackslash{} 2023.md | sed -n `1,120p'}\\
\textbf{Tool response:} (...61 lines omitted...)\\
\\
\textbf{Tool call:}
\texttt{nl -ba /files/session9\_11:22\textbackslash{} am\textbackslash{} on\textbackslash{} 13\textbackslash{} March,\textbackslash{} 2023.md | sed -n `1,120p'}\\
\textbf{Tool response:} (...66 lines omitted...)\\
\\
\textbf{Tool call:}
\texttt{nl -ba /files/session22\_5:33\textbackslash{} pm\textbackslash{} on\textbackslash{} 26\textbackslash{} August,\textbackslash{} 2023.md | sed -n `1,140p'}\\
\textbf{Tool response:} (...106 lines omitted...)\\
\\
\textbf{Assistant}:
The year is not specified. Jolene confirms she practices yoga ("I've been into yoga and meditation lately") but no conversation entry gives a specific year when she began. Another transcript notes "That was when I first started doing yoga," yet it also does not state a year. Therefore the files do not specify the year Jolene started yoga.
\end{tcolorbox}

\end{table}

\begin{table}
\caption{\label{tab:locomo-expindex}
\sys{} on the same Locomo example as \Cref{tab:locomo-react}.
\textcolor{ForestGreen}{Green text refers to changes due to \sys{}.}
}
\begin{tcolorbox}[title=System prompt (\sys{})]
\textcolor{ForestGreen}{
You are given a set of files with past experiences, each containing a file path, past user questions, descriptions of how the file was used to answer them, and the specific snippets that were used.\\
\\
When handling a new task, you should:\\
(1) Consult the experiences as prior examples. Use them to guide your reasoning, retrieval strategy, and expectations about how files may be relevant.\\
(2) Determine relevance. Identify which, if any, of the past experiences apply to the new task. If none are relevant, continue normally without forcing a connection.\\
(3) Always verify against the current files. The experiences are helpful context, but the actual files available now take priority. If there is any discrepancy, updated or current file contents override the past experiences.\\
\\
Past experiences:\\
<file file\_path='/files/session13\_3:56 pm on 6 June, 2023.md'>\\
<experience>\\
Question:\\
How long has Jolene been doing yoga?\\
\\
Contribution summary:\\
The cited file (/files/session13\_3:56 pm on 6 June, 2023.md) contains the transcript line where Jolene says, "I've been doing them sporadically for about 3 years now" (line 67). That direct quote from the session serves as the primary evidence supporting the answer that she has been doing yoga for about three years.\\
\\
Snippets:\\
line 67: I've been doing them sporadically for about 3 years now and they've had a real positive effect on me.\\
</experience>\\
</file>
}
\end{tcolorbox}

\begin{tcolorbox}[title=Task]
Which year did Jolene start practicing yoga?
\end{tcolorbox}

\begin{tcolorbox}[title=Assistant messages]
\textcolor{ForestGreen}{
\textbf{Tool call:}
\texttt{sed -n `50,70p' /files/session13\_3:56\textbackslash{} pm\textbackslash{} on\textbackslash{} 6\textbackslash{} June,\textbackslash{} 2023.md:53}\\
\textbf{Tool response:}\\
I've been doing them sporadically for about 3 years now and they've had a real positive effect on me.\\
(...19 lines omitted...)\\
\\
\textbf{Assistant}: The answer is 2020. She said she's been doing yoga “sporadically for about 3 years now”, which places the start at about 2020.
}
\end{tcolorbox}

\end{table}

\clearpage

\subsection{Artifact-pair experience}

\Cref{tab:mmqa} illustrates an example where \sys{} leverages artifact-pair experience.

\begin{table}[h!]
\caption{\label{tab:mmqa}On MMQA, the result becomes correct once the new tables introduced through edge connections by \sys{} are incorporated. \textcolor{ForestGreen}{Green text refers to changes due to \sys{}.}}

\begin{tcolorbox}[title=Task]
How many distinct orders that are currently `On Road' have at least one `music' product item with a status of `Payed'?
\end{tcolorbox}
\begin{tcolorbox}[title=Assistant messages]
\textbf{Tool call:} \texttt{vector\_search(query: "orders On Road music product Payed distinct orders 'On Road' 'music' 'Payed'")}\\
\\
\textbf{Tool response:}\\
(...3 tables omitted ...)\\
Table: Order\_Items\_d9a3...\\

Table: Products\_46df...\\
Sample rows:\\
|   product\_id | color\_code   | product\_category\_code   | product\_name   | typical\_buying\_price   | typical\_selling\_price   | product\_description   | other\_product\_details   |\\
|-------------:|:-------------|:------------------------|:---------------|:-----------------------|:------------------------|:----------------------|:------------------------|\\
|            1 | '4'          | 'Spices'                | 'cumin'        | ''                     | '2878.3'                | 'et'                  | None                    |\\
|            2 | '2'          | 'Spices'                | 'peper'        | '352447.2874677'       | '1892070.2803543'       | 'rerum'               | None                    |\\

\textcolor{ForestGreen}{
Table: Products\_eafb...\\
Sample rows:
|   product\_id | product\_name   | product\_details   |\\
|-------------:|:---------------|:------------------|\\
|            1 | 'food'         | None              |\\
|            2 | 'book'         | None              |\\
(Introduced via a relationship with \texttt{Order\_Items\_d9a3}:
\texttt{Order\_Items\_d9a3.product\_id = Products\_eafb.product\_id})}
\\

\textbf{Assistant}:\\
ReAct: I will filter ``music'' product using \texttt{Products\_46df.product\_category\_code = `music'}...

\textcolor{ForestGreen}{ReAct with \sys{}: I wonder if `music' might be listed as a product name in a products table...I'll use SQL to search both \texttt{Products\_46df} and \texttt{Products\_eafb} to see what I can find...}
\end{tcolorbox}
\end{table}

\clearpage

\section{Evaluation details}
\label{app:eval}

For each dataset, we sample evaluation tasks from the full collection to include a set of roughly 100 tasks.

For Swebench-verified, we report the answer score as the percentage of issues successfully resolved, following the evaluation protocol in~\citet{jimenez2023swe}.

For Spider, we use execution accuracy~\citep{yu2018spider}, assigning a score of 1 when the predicted SQL query yields the same output as the ground-truth query and 0 otherwise.

For MMQA, Musique, and Locomo, the answer score is based on LLM‑judged correctness, comparing the predicted answer with the ground-truth answer. For MMQA and Musique, we adopt the judge prompt from~\citet{li20262}. For Locomo, we use the prompt from~\citet{chhikara2025mem0}.

For SQA and Freshstack, we compute the answer score using an LLM‑judged rubric-based score (normalized to the range 0-1), comparing the predicted answer with human‑designed rubrics. For SQA, we use the prompt from~\citet{bragg2025astabench}. For Freshstack, we use the prompt from~\citet{chang2026karl}.

Following the setup in~\citet{bragg2025astabench}, we use \texttt{Gemini‑2.5‑Flash} as the judge LLM for all datasets requiring LLM‑judge‑based scoring.

